\documentclass[final]{cvpr}
\usepackage[pagebackref=true,breaklinks=true,colorlinks,bookmarks=false]{hyperref}

\usepackage{times}
\usepackage{graphicx}
\usepackage{amsmath,amssymb,amsthm,mathabx}
\usepackage{booktabs}
\usepackage{acronym}
\usepackage{enumitem}
\usepackage{balance}
\usepackage{setspace}
\usepackage[dvipsnames]{xcolor}
\usepackage[capitalise]{cleveref}
\usepackage{tabularx,multirow,array,makecell}
\makeatletter
\DeclareRobustCommand\onedot{\futurelet\@let@token\@onedot}
\def\@onedot{\ifx\@let@token.\else.\null\fi\xspace}

\def\ie{\emph{i.e}\onedot}

\def\etal{\emph{et al}\onedot}
\makeatother

\newcommand\blfootnote[1]{%
  \begingroup
  \renewcommand\thefootnote{}\footnote{#1}%
  \addtocounter{footnote}{-1}%
  \endgroup
}

\makeatletter
\renewcommand{\paragraph}{%
  \@startsection{paragraph}{4}%
  {\z@}{0ex \@plus 0ex \@minus 0ex}{-1em}%
  {\hskip\parindent\normalfont\normalsize\bfseries}%
}
\makeatother

\graphicspath{{figures/}}

\crefname{algocf}{Alg.}{Algs.}
\Crefname{algocf}{Algorithm}{Algorithms}

\definecolor{gblue}{HTML}{4285F4}
\definecolor{gred}{HTML}{DB4437}

\acrodef{vqa}[VQA]{Visual Question Answering}
\acrodef{rpm}[RPM]{Raven's Progressive Matrices}
\acrodef{wren}[WReN]{Wild Relational Network}
\acrodef{ai}[AI]{Artificial Intelligence}
\acrodef{jsd}[JSD]{Jensen–Shannon Divergence}
\acrodef{prae}[PrAE]{Probabilistic Abduction and Execution}

\def\cvprPaperID{xxxx} % *** Enter the CVPR Paper ID here
\def\confYear{CVPR 2021}

\begin{document}
\title{Abstract Spatial-Temporal Reasoning via Probabilistic Abduction and Execution}

\author{Chi Zhang$^{\star}$ \qquad Baoxiong Jia$^{\star}$ \qquad Song-Chun Zhu \qquad Yixin Zhu \\
UCLA Center for Vision, Cognition, Learning, and Autonomy \\
{\tt\small \{chi.zhang,baoxiongjia\}@ucla.edu, sczhu@stat.ucla.edu, yixin.zhu@ucla.edu}
% For a paper whose authors are all at the same institution,
% omit the following lines up until the closing ``}''.
% Additional authors and addresses can be added with ``\and'',
% just like the second author.
% To save space, use either the email address or home page, not both
}

\maketitle
\thispagestyle{empty}
\pagestyle{empty}
\blfootnote{$^\star$ indicates equal contribution.}

\begin{abstract}
Spatial-temporal reasoning is a challenging task in \acf{ai} due to its demanding but unique nature: a theoretic requirement on \emph{representing} and \emph{reasoning} based on spatial-temporal knowledge in mind, and an applied requirement on a high-level cognitive system capable of \emph{navigating} and \emph{acting} in space and time. Recent works have focused on an abstract reasoning task of this kind---\acf{rpm}. Despite the encouraging progress on \ac{rpm} that achieves human-level performance in terms of accuracy, modern approaches have neither a treatment of human-like reasoning on generalization, nor a potential to generate answers. To fill in this gap, we propose a neuro-symbolic \textbf{\acf{prae}} learner; central to the \ac{prae} learner is the process of probabilistic abduction and execution on a probabilistic scene representation, akin to the mental manipulation of objects. Specifically, we disentangle perception and reasoning from a monolithic model. The neural visual perception frontend predicts objects' attributes, later aggregated by a scene inference engine to produce a probabilistic scene representation. In the symbolic logical reasoning backend, the \ac{prae} learner uses the representation to \textbf{abduce} the hidden rules. An answer is predicted by \textbf{executing} the rules on the probabilistic representation. The entire system is trained end-to-end in an analysis-by-synthesis manner \textbf{without} any visual attribute annotations. Extensive experiments demonstrate that the \ac{prae} learner improves cross-configuration generalization and is capable of rendering an answer, in contrast to prior works that merely make a categorical choice from candidates.
\end{abstract}

\section{Introduction}\label{sec:intro}

While ``thinking in pictures''~\cite{grandin2006thinking}, \ie, spatial-temporal reasoning, is effortless and instantaneous for humans, this significant ability has proven to be particularly challenging for current machine vision systems~\cite{jang2017tgif}. With the promising results~\cite{grandin2006thinking} that show the very ability is strongly correlated with one's logical induction performance and a crucial factor for the intellectual history of technology development, recent computational studies on the problem focus on an abstract reasoning task relying heavily on ``thinking in pictures''---\acf{rpm}~\cite{carpenter1990one,hunt1974quote,raven1936mental,raven1998raven}. In this task, a subject is asked to pick a correct answer that best fits an incomplete figure matrix to satisfy the hidden governing rules. The ability to solve \ac{rpm}-like problems is believed to be critical for generating and conceptualizing solutions to multi-step problems, which requires mental manipulation of given images over a time-ordered sequence of spatial transformations. Such a task is also believed to be characteristic of relational and analogical reasoning and an indicator of one's \emph{fluid intelligence}~\cite{snow1984the,hofstadter1995fluid,jaeggi2008improving,spearman1927abilities}. 

State-of-the-art algorithms incorporating a contrasting mechanism and perceptual inference~\cite{hill2019learning,zhang2019learning} have achieved decent performance in terms of accuracy. Nevertheless, along with the improved accuracy from deep models come critiques on its transparency, interpretability, generalization, and difficulty to incorporate knowledge. Without explicitly distinguishing perception and reasoning, existing methods use a \textit{monolithic} model to learn correlation, sacrificing transparency and interpretability in exchange for improved performance~\cite{hill2019learning,hu2021stratified,santoro2018measuring,wang2020abstract,zhang2019raven,zhang2019learning,zheng2019abstract}. Furthermore, as shown in experiments, deep models nearly always overfit to the training regime and cannot properly generalize. Such a finding is consistent with Fodor~\cite{fodor1988connectionism} and Marcus's~\cite{marcus1998rethinking,marcus2018algebraic} hypothesis that human-level systematic generalizability is hardly compatible with classic neural networks; Marcus postulates that a neuro-symbolic architecture should be recruited for human-level generalization~\cite{edmonds2018human,edmonds2020theory,edmonds2019decomposing,marcus2019rebooting,marcus2020insights,xie2021halma}.

Another defect of prior methods is the lack of top-down and bottom-up reasoning~\cite{zhang2019learning}: Human reasoning applies a \textit{generative} process to abduce rules and execute them to synthesize a possible solution in mind, and \textit{discriminatively} selects the most similar answer from choices~\cite{holyoak2012oxford}. This bi-directional reasoning is in stark contrast to discriminative-only models, solely capable of making a categorical choice.

Psychologists also call for weak attribute supervision in \ac{rpm}. As isolated Amazonians, absent of schooling on primitive attributes, could still correctly solve \ac{rpm}~\cite{dehaene2006core,izard2011flexible}, an ideal computational counterpart should be able to learn it \emph{absent of visual attribute annotations}. This weakly-supervised setting introduces unique challenges: How to jointly learn these visual attributes given only ground-truth images? With uncertainties in perception, how to abduce hidden logic relations from it? How about executing the symbolic logic on inaccurate perception to derive answers?
 
\begin{figure}[t!]
    \centering
    \includegraphics[width=\linewidth]{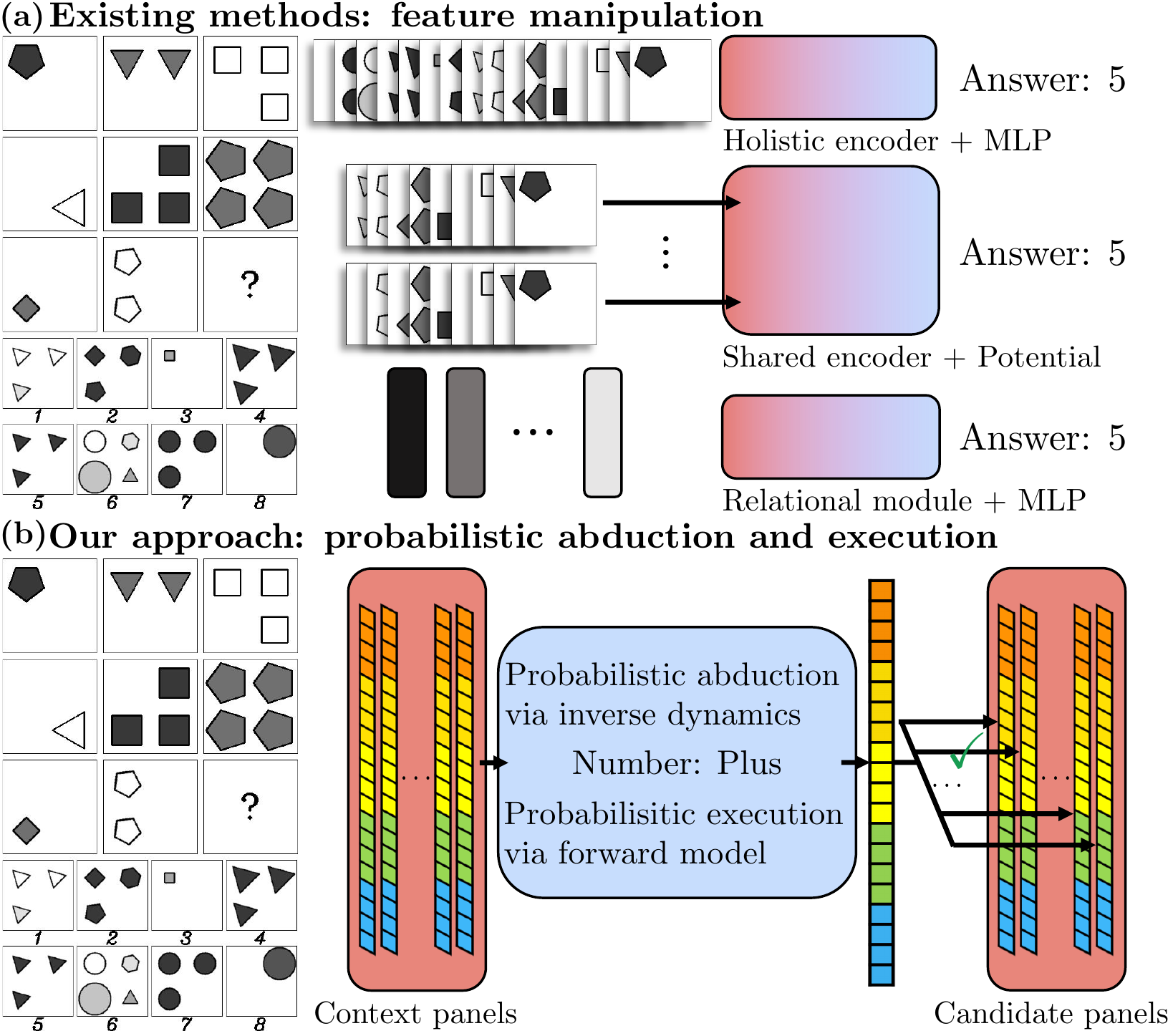}
    \caption{Differences between (a) prior methods and (b) the proposed approach. Prior methods do not explicitly distinguish {\color{gred}perception} and {\color{gblue}reasoning}; instead, they use a monolithic model and only differ in how features are manipulated, lacking semantics and probabilistic interpretability. In contrast, the proposed approach disentangles this monolithic process: It {\color{gred}perceives} each panel of \ac{rpm} as a set of probability distributions of attributes, performs logical {\color{gblue}reasoning} to abduce the hidden rules that govern the time-ordered sequence, and executes the abduced rules to \emph{generate} answer representations. A final choice is made based on the divergence between predicted answer distributions and each candidate's distributions; see \cref{sec:related} for a detailed comparison.}
    \label{fig:comparison}
    \vspace{-20pt}
\end{figure}

To support cross-configuration generalization and answer generation, we move a step further towards a neuro-symbolic model with explicit logical reasoning and human-like generative problem-solving while addressing the challenges. Specifically, we propose the \emph{\acf{prae}} learner; central to it is the process of abduction and execution on the probabilistic scene representation. Inspired by Fodor, Marcus, and neuro-symbolic reasoning~\cite{han2019visual,mao2019neuro,yi2020clevrer,yi2018neural}, the \ac{prae} learner disentangles the previous monolithic process into two separate modules: a neural visual perception frontend and a symbolic logical reasoning backend. The neural visual frontend operates on object-based representation~\cite{han2019visual,kansky2017schema,mao2019neuro,yi2020clevrer,yi2018neural} and predicts conditional probability distributions on its attributes. A scene inference engine then aggregates all object attribute distributions to produce a probabilistic scene representation for the backend. The symbolic logical backend abduces, from the representation, hidden rules that govern the time-ordered sequence via inverse dynamics. An execution engine executes the rules to \emph{generate} an answer representation in a probabilistic planning manner~\cite{ghallab2004automated,huang2019continuous,konidaris2015symbol}, instead of directly making a categorical choice among the candidates. The final choice is selected based on the divergence between the generated prediction and the given candidates. The entire system is trained end-to-end with a cross-entropy loss and a curricular auxiliary loss~\cite{santoro2018measuring,zhang2019raven,zhang2019learning} \emph{without} any visual attribute annotations. \cref{fig:comparison} compares the proposed \ac{prae} learner with prior methods.

The unique design in \ac{prae} connects perception and reasoning and offers several advantages: 
(i) With an intermediate probabilistic scene representation, the neural visual perception frontend and the symbolic logical reasoning backend can be \emph{swapped} for different task domains, enabling a greater extent of module reuse and combinatorial \emph{generalization}.
(ii) Instead of blending perception and reasoning into one monolithic model without any explicit reasoning, probabilistic abduction offers a more \emph{interpretable} account for reasoning on a logical representation. It also affords a more detailed analysis into both perception and reasoning.
(iii) Probabilistic execution permits a \emph{generative} process to be integrated into the system. Symbolic logical constraints can be transformed by the execution engine into a forward model~\cite{jordan1992forward} and applied in a probabilistic manner to predict the final scene representation, such that the entire system can be trained by analysis-by-synthesis~\cite{chen2019holistic,grenander1976lectures,han2019divergence,huang2018cooperative,huang2018holistic,loper2014opendr,wu2017neural,wu2017marrnet,xie2016theory,xie2019learning,yuille2006vision,zhu1998filters}.
(iv) Instead of making a deterministic decision or drawing limited samples, maintaining probabilistic distributions brings in extra robustness and fault tolerance and allows gradients to be easily propagated.

This paper makes three major contributions: 
(i) We propose the \emph{\acf{prae}} learner. Unlike previous methods, the \ac{prae} learner disentangles perception and reasoning from a monolithic model with the reasoning process realized by abduction and execution on a probabilistic scene representation. The abduction process performs interpretable reasoning on perception results. The execution process adds to the learner a generative flavor, such that the system can be trained in an analysis-by-synthesis manner without any visual attribute annotations. 
(ii) Our experiments demonstrate the \ac{prae} learner achieves better generalization results compared to existing methods in the cross-configuration generalization task of \ac{rpm}. We also show that the \ac{prae} learner is capable of generating answers for \ac{rpm} questions via a renderer. 
(iii) We present analyses into the inner functioning of both perception and reasoning, providing an interpretable account of \ac{prae}.

\section{Related Work}\label{sec:related}

\paragraph{Neuro-Symbolic Visual Reasoning} 

Neuro-symbolic methods have shown promising potential in tasks involving an interplay between vision and language and vision and causality. Qi \etal~\cite{qi2020generalized,qi2018generalized} showed that action recognition could be significantly improved with the help of grammar parsing, and Li \etal~\cite{li2020closed} integrated perception, parsing, and logics into a unified framework. Of particular relevance, Yi \etal~\cite{yi2018neural} first demonstrated a prototype of a neuro-symbolic system to solve \ac{vqa}~\cite{antol2015vqa}, where the vision system and the language parsing system were separately trained with a final symbolic logic system applying the parsed program to deliver an answer. Mao \etal~\cite{mao2019neuro} improved such a system by making the symbolic component continuous and end-to-end trainable, despite sacrificing the semantics and interpretability of logics. Han \etal~\cite{han2019visual} built on~\cite{mao2019neuro} and studied the metaconcept problem by learning concept embeddings. A recent work investigated temporal and causal relations in collision events~\cite{yi2020clevrer} and solved it in a way similar to~\cite{yi2018neural}. The proposed \ac{prae} learner is similar to but has fundamental differences from existing neuro-symbolic methods. Unlike the method proposed by Yi \etal~\cite{yi2020clevrer,yi2018neural}, our approach is end-to-end trainable and does not require intermediate visual annotations, such as ground-truth attributes. Compared to \cite{mao2019neuro}, our approach preserves logic semantics and interpretability by explicit logical reasoning involving probabilistic abduction and execution in a probabilistic planning manner~\cite{ghallab2004automated,huang2019continuous,konidaris2015symbol}.

\paragraph{Computational Approaches to \ac{rpm}} 

Initially proposed as an intelligence quotient test into general intelligence and fluid intelligence~\cite{raven1936mental,raven1998raven}, \acf{rpm} has received notable attention from the research community of cognitive science. Psychologists have proposed reasoning systems based on symbolic representations and discrete logics~\cite{carpenter1990one,lovett2017modeling,lovett2010structure,lovett2009solving}. However, such logical systems cannot handle visual uncertainty arising from imperfect perception. Similar issues also pose challenges to methods based on image similarity~\cite{little2012bayesian,mcgreggor2014confident,mcgreggor2014fractals,mekik2018similarity,shegheva2018structural}. Recent works approach this problem in a data-driven manner. The first automatic \ac{rpm} generation method was proposed by Wang and Su~\cite{wang2015automatic}. Santoro \etal~\cite{santoro2018measuring} extended it using procedural generation and introduced the \ac{wren} to solve the problem. Zhang \etal~\cite{zhang2019raven} and Hu \etal~\cite{hu2021stratified} used stochastic image grammar~\cite{zhu2007stochastic} and provided structural annotations to the dataset. Unanimously, existing methods do not explicitly distinguish perception and reasoning; instead, they use one monolithic neural model, sacrificing interpretability in exchange for better performance. The differences in previous methods lie in how features are manipulated: Santoro \etal~\cite{santoro2018measuring} used the relational module to extract final features, Zhang \etal~\cite{zhang2019raven} stacked all panels into the channel dimension and fed them into a residual network, Hill \etal~\cite{hill2019learning} prepared the data in a contrasting manner, Zhang \etal~\cite{zhang2019learning} composed the context with each candidate and compared their potentials, Wang \etal~\cite{wang2020abstract} modeled the features by a multiplex graph, and Hu \etal~\cite{hu2021stratified} integrated hierarchical features. Zheng \etal~\cite{zheng2019abstract} studied a teacher-student setting in \ac{rpm}, while Steenbrugge \etal~\cite{steenbrugge2018improving} focused on a generative approach to improve learning. Concurrent to our work, Spratley \etal~\cite{spratley2020closer} unsupervisedly extracted object embeddings and conducted reasoning via a ResNet. In contrast, \ac{prae} is designed to address cross-configuration generalization and disentangles perception and reasoning from a monolithic model, with symbolic logical reasoning implemented as probabilistic abduction and execution.

\section{The PrAE Learner}\label{sec:method}

\begin{figure*}[t!]
    \centering
    \includegraphics[width=\linewidth]{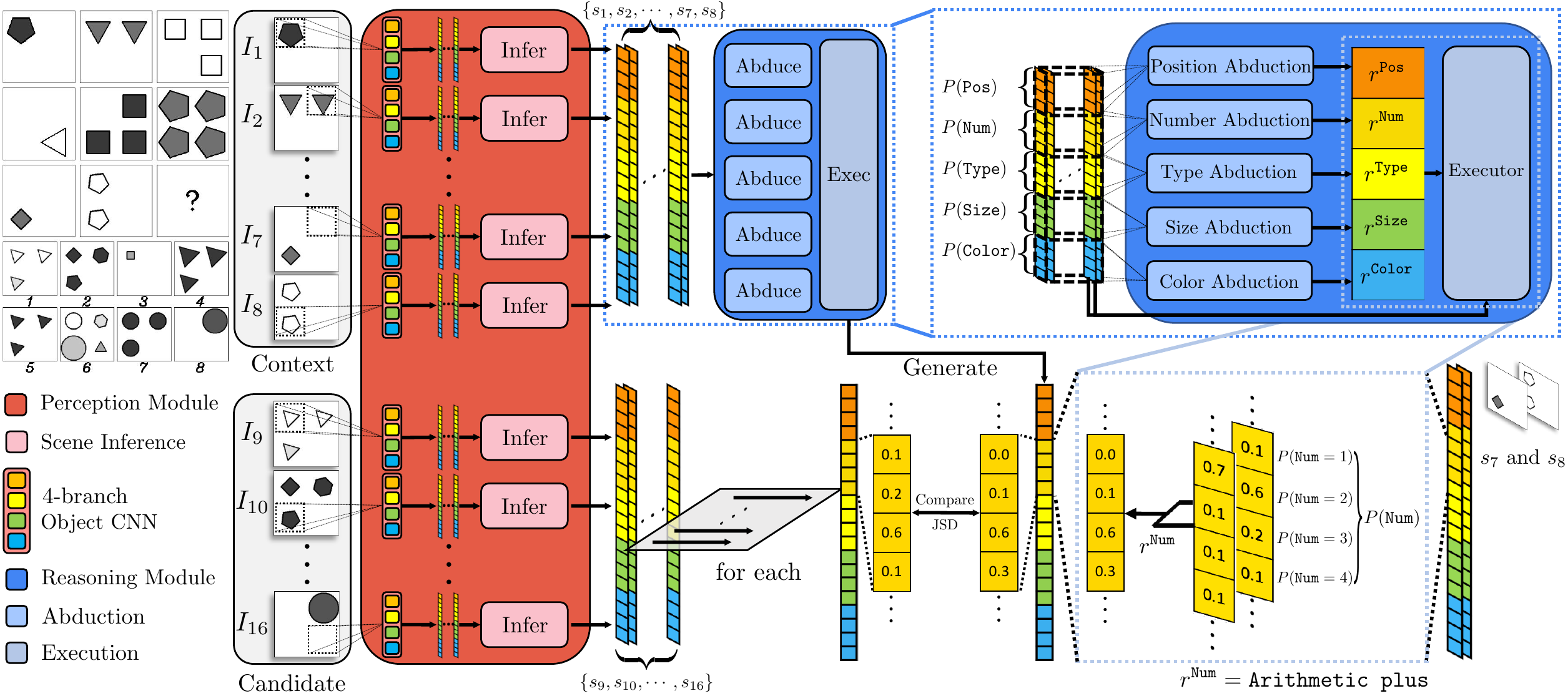}
    \caption{An overview of learning and reasoning of the proposed \ac{prae} learner. Given an \ac{rpm} instance, the neural perception frontend (in red) extracts probabilistic scene representation for each of the 16 panels (8 contexts + 8 candidates). The \emph{Object CNN} sub-module takes in each image region returned by a sliding window to produce object attribute distributions (over objectiveness, type, size, and color). The \emph{Scene Inference Engine} sub-module (in pink) aggregates object attribute distributions from all regions to produce panel attribute distributions (over position, number, type, size, and color). Probabilistic representation for context panels is fed into the symbolic reasoning backend (in blue), which abduces hidden rule distributions for all panel attributes (upper-right figure) and executes chosen rules on corresponding context panels to generate the answer representation (lower-right figure). The answer representation is compared with each candidate representation from the perception frontend; the candidate with minimum divergence from the prediction is chosen as the final answer. The lower-right figure is an example of probabilistic execution on the panel attribute of $\mathtt{Number}$; see \cref{sec:reasoning} for the exact computation process.}
    \label{fig:overview}
\end{figure*}

\paragraph{Problem Setup} 

In this section, we explain our approach to tackling the \ac{rpm} problem. Each \ac{rpm} instance consists of $16$ panels: $8$ context panels form an incomplete $3\times3$ matrix with a $9$th missing entry, and $8$ candidate panels for one to choose. The goal is to pick one candidate that best completes the matrix to satisfy the latent governing rules. Existing datasets~\cite{hu2021stratified,santoro2018measuring,wang2015automatic,zhang2019raven} assume fixed sets of object attributes, panel attributes, and rules, with each panel attribute governed by one rule. The value of a panel attribute constrains the value of the corresponding object attribute for each object in it.

\paragraph{Overview} 

The proposed neuro-symbolic \ac{prae} learner disentangles previous monolithic visual reasoning into two modules: the neural visual perception frontend and the symbolic logical reasoning backend. The frontend uses a CNN to extract object attribute distributions, later aggregated by a scene inference engine to produce panel attribute distributions. The set of all panel attribute distributions in a panel is referred to as its \emph{probabilistic scene representation}. The backend retrieves this compact scene representation and performs logical abduction and execution in order to predict the answer representation in a generative manner. A final choice is made based on the divergence between the prediction and each candidate. Using REINFORCE~\cite{williams1992simple}, the entire system is trained \emph{without attribute annotations} in a curricular manner; see \cref{fig:overview} for an overview of \ac{prae}.

\subsection{Neural Visual Perception}

The neural visual perception frontend operates on each of the $16$ panels \emph{independently} to produce probabilistic scene representation. It has two sub-modules: object CNN and scene inference engine.

\paragraph{Object CNN} 

Given an image panel $I$, a sliding window traverses its spatial domain and feeds each image region into a $4$-branch CNN. The $4$ CNN branches use the same LeNet-like architecture~\cite{lecun1998gradient} and produce the probability distributions of object attributes, including objectiveness (whether the image region has an object), type, size, and color. Of note, the distributions of type, size, and color are conditioned on objectiveness being true. Attribute distributions of each image region are kept and sent to the scene inference engine to produce panel attribute distributions.

\paragraph{Scene Inference Engine}

The scene inference engine takes in the outputs of object CNN and produces panel attribute distributions (over position, number, type, size, and color) by marginalizing over the set of object attribute distributions (over objectiveness, type, size, and color). Take the panel attribute of $\mathtt{Number}$ as an example: Given $N$ objectiveness probability distributions produced by the object CNN for $N$ image regions, the probability of a panel having $k$ objects can be computed as
\begin{equation}
    P(\mathtt{Number} = k) = \sum_{\substack{B^o \in \{0, 1\}^N \\ |B^o| = k}} \prod_{j=1}^N P(b_j^o = B_j^o),
    \label{eqn:number_eg}
\end{equation}
where $B^o$ is an ordered binary sequence corresponding to objectiveness of the $N$ regions, $|\cdot|$ the number of $1$ in the sequence, and $P(b_j^o)$ the objectiveness distribution of the $j$th region. We assume $k \geq 1$ in each \ac{rpm} panel, leave $P(\mathtt{Number}=0)$ out, and renormalize the probability to have a sum of $1$. The panel attribute distributions for position, type, size, and color, can be computed similarly. 

We refer to the set of all panel attribute distributions in a panel its \emph{probabilistic scene representation}, denoted as $s$, with the distribution of panel attribute $a$ denoted as $P(s^a)$. 

\subsection{Symbolic Logical Reasoning}\label{sec:reasoning}

The symbolic logical reasoning backend collects probabilistic scene representation from $8$ context panels, abduces the probability distributions over hidden rules on each panel attribute, and executes them on corresponding panels of the context. Based on a prior study~\cite{carpenter1990one}, we assume a set of symbolic logical constraints describing rules is available. For example, the $\mathtt{Arithmetic}$ $\mathtt{plus}$ rule on $\mathtt{Number}$ can be represented as: for each row (column), $\forall l, m \geq 1$
\begin{equation}\small
    (\mathtt{Number}_1 = m) \land (\mathtt{Number}_2 = l) \land (\mathtt{Number}_3 = m + l),%
\end{equation}
where $\mathtt{Number}_i$ denotes the number of objects in the $i$th panel in a row (column). With access to such constraints, we use inverse dynamics to abduce the rules in an instance. They can also be transformed into a forward model and executed on discrete symbols: For instance, $\mathtt{Arithmetic}$ $\mathtt{plus}$ deterministically adds $\mathtt{Number}$ in the first two panels to obtain the $\mathtt{Number}$ of the last panel.

\paragraph{Probabilistic Abduction} 

Given the probabilistic scene representation of $8$ context panels, the probabilistic abduction engine calculates the probability of rules for each panel attribute via inverse dynamics. Formally, for each rule $r$ on a panel attribute $a$,
\begin{equation}
    P(r^a \mid I_1, \ldots, I_8) = P(r^a \mid I_1^a, \ldots, I_8^a),
    \label{eqn:gen_inv_dyn}
\end{equation}
where $I_i$ denotes the $i$th context panel, and $I_i^a$ the component of context panel $I_i$ corresponding to $a$. Note \cref{eqn:gen_inv_dyn} generalizes inverse dynamics~\cite{jordan1992forward} to $8$ states, in contrast to that of a conventional MDP.

To model $P(r^a \mid I_1^a, \ldots, I_8^a)$, we leverage the compact probabilistic scene representation with respect to attribute $a$ and logical constraints:
\begin{equation}
    P(r^a \mid I_1^a, \ldots, I_8^a) \propto \sum_{S^a \in \mathtt{valid}(r^a)} \prod_{i = 1}^8 P(s_i^a = S_i^a),
\end{equation}
where $\mathtt{valid}(\cdot)$ returns a set of attribute value assignments of the context panels that satisfy the logical constraints of $r^a$, and $i$ indexes into context panels. By going over all panel attributes, we have the distribution of hidden rules for each of them.

Take $\mathtt{Arithmetic}$ $\mathtt{plus}$ on $\mathtt{Number}$ as an example. A row-major assignment for context panels can be $[1, 2, 3, 1, 3, 4, 1, 2]$ (as in \cref{fig:overview}), whose probability is computed as the product of each panel having $k$ objects as in \cref{eqn:number_eg}. Summing it with other assignment probabilities gives an unnormalized rule probability.

We note that the set of valid states for each $r^a$ is a product space of valid states on each row (column). Therefore, we can perform partial marginalization on each row (column) first and aggregate them later to avoid directly marginalizing over the entire space. This decomposition will help reduce computation and mitigate numerical instability.

\paragraph{Probabilistic Execution} 

For each panel attribute $a$, the probabilistic execution engine chooses a rule from the abduced rule distribution and executes it on corresponding context panels to predict, in a generative fashion, the panel attribute distribution of an answer. While traditionally, a logical forward model only works on discrete symbols, we follow a generalized notion of probabilistic execution as done in probabilistic planning~\cite{huang2019continuous,konidaris2015symbol}. The probabilistic execution could be treated as a distribution transformation that redistributes the probability mass based on logical rules. For a binary rule $r$ on $a$,
\begin{equation}\small
    P(s_3^a = S_3^a) \propto \sum_{\substack{(S_2^a, S_1^a) \in \mathtt{pre}(r^a) \\ S_3^a = f(S_2^a, S_1^a; r^a)}} P(s_2^a = S_2^a) P(s_1^a = S_1^a),
    \label{eqn:execute}
\end{equation}
where $f$ is the forward model transformed from logical constraints and $\mathtt{pre}(\cdot)$ the rule precondition set. Predicted distributions of panel attributes compose the final probabilistic scene representation $s_f$.

As an example of $\mathtt{Arithmetic}$ $\mathtt{plus}$ on $\mathtt{Number}$, $4$ objects result from the addition of $(1, 3)$, $(2, 2)$, and $(3, 1)$. The probability of an answer having $4$ objects is the sum of the instances' probabilities.

During training, the execution engine samples a rule from the abduced probability. During testing, the most probable rule is chosen.

\paragraph{Candidate Selection} 

With a set of predicted panel attribute distributions, we compare it with that from each candidate answer. We use the \ac{jsd}~\cite{lin1991divergence} to quantify the divergence between the prediction and the candidate, \ie, 
\begin{equation}
    d(s_f, s_i) = \sum_a \mathbb{D}_{\text{JSD}}(P(s_f^a) \mid\mid P(s_i^a)),
    \label{eqn:dist}
\end{equation}
where the summation is over panel attributes and $i$ indexes into the candidate panels. The candidate with minimum divergence will be chosen as the final answer.

\paragraph{Discussion} 

The design of reasoning as probabilistic abduction and execution is a computational and interpretable counterpart to human-like reasoning in \ac{rpm}~\cite{carpenter1990one}. By abduction, one infers the hidden rules from context panels. By executing the abduced rules, one obtains a probabilistic answer representation. Such a probabilistic representation is compared with all candidates available; the most similar one in terms of divergence is picked as the final answer. Note that the probabilistic execution adds the generative flavor into reasoning: \cref{eqn:execute} depicts the predicted panel attribute distribution, which can be sampled and sent to a rendering engine for panel generation. The entire process resembles bi-directional inference and combines both top-down and bottom-up reasoning missing in prior works. In the meantime, the design addresses challenges mentioned in \cref{sec:intro} by marginalizing over perception and abducing and executing rules probabilistically.

\subsection{Learning Objective}

During training, we transform the divergence in \cref{eqn:dist} into a probability distribution by
\begin{equation}
    P(\text{Answer} = i) \propto \exp(-d(s_f, s_i))
\end{equation}
and minimize the cross-entropy loss. Note that the learning procedure follows a general paradigm of analysis-by-synthesis~\cite{chen2019holistic,grenander1976lectures,han2019divergence,huang2018cooperative,huang2018holistic,loper2014opendr,wu2017neural,wu2017marrnet,xie2016theory,xie2019learning,yuille2006vision,zhu1998filters}: The learner synthesizes a result and measures difference analytically.

As the reasoning process involves rule selection, we use REINFORCE~\cite{williams1992simple} to optimize:
\begin{equation}
    \underset{\theta}{\text{min}}\ \mathbb{E}_{P(r)}[\ell(P(\text{Answer}; r), y)],
\end{equation}
where $\theta$ denotes the trainable parameters in the object CNN, $P(r)$ packs the rule distributions over all panel attributes, $\ell$ is the cross-entropy loss, and $y$ is the ground-truth answer. Note that here we make explicit the dependency of the answer distribution on rules, as the predicted probabilistic scene representation $s_f$ is dependent on the rules chosen.

In practice, the \ac{prae} learner experiences difficulty in convergence with cross-entropy loss only, as the object CNN fails to produce meaningful object attribute predictions at the early stage of training. To resolve this issue, we jointly train the \ac{prae} learner to optimize the auxiliary loss, as discussed in recent literature~\cite{santoro2018measuring,zhang2019raven,zhang2019learning}. The auxiliary loss regularizes the perception module such that the learner produces the correct rule prediction. The final objective is
\begin{equation}
    \underset{\theta}{\text{min}}\ \mathbb{E}_{P(r)}[\ell(P(\text{Answer}; r), y)] + \sum_a \lambda^a \ell(P(r^a), y^a),
\end{equation}
where $\lambda^a$ is the weight coefficient, $P(r^a)$ the distribution of the abduced rule on $a$, and $y^a$ the ground-truth rule. In reinforcement learning terminology, one can treat the cross-entropy loss as the negative reward and the auxiliary loss as behavior cloning~\cite{sutton1998introduction}.

\subsection{Curriculum Learning}

In preliminary experiments, we notice that accurate objectiveness prediction at the early stage is essential to the success of the learner, while learning without auxiliary will reinforce the perception system to produce more accurate object attribute predictions in the later stage when all branches of the object CNN are already warm-started. This observation is consistent with human learning: One learns object attributes only after they can correctly distinguish objects from the scene, and their perception will be enhanced with positive signals from the task.

Based on this observation, we train our \ac{prae} learner in a $3$-stage curriculum~\cite{bengio2009curriculum}. In the first stage, only parameters corresponding to objectiveness are trained. In the second stage, objectiveness parameters are frozen while weights responsible for type, size, and color prediction are learned. In the third stage, we perform joint fine-tuning for the entire model via REINFORCE~\cite{williams1992simple}.

\begin{table*}[t!]
    \centering
    \resizebox{\linewidth}{!}{
    \begin{tabular}{l c c c c c c c c}
        \toprule
        Method                & Acc             & Center          & 2x2Grid         & 3x3Grid         & L-R             & U-D             & O-IC            & O-IG           \\
        \midrule
        \ac{wren}             & $9.86 / 14.87$  & $8.65 / 14.25$  & $29.60 / 20.50$ & $9.75 / 15.70$  & $4.40 / 13.75$  & $5.00 / 13.50$  & $5.70 / 14.15$  & $5.90 / 12.25$ \\
        LSTM                  & $12.81 / 12.52$ & $12.70 / 12.55$ & $13.80 / 13.50$ & $12.90 / 11.35$ & $12.40 / 14.30$ & $12.10 / 11.35$ & $12.45 / 11.55$ & $13.30 / 13.05$ \\
        LEN                   & $12.29 / 13.60$ & $11.85 / 14.85$ & $41.40 / 18.20$ & $12.95 / 13.35$ & $3.95 / 12.55$  & $3.95 / 12.75$  & $5.55 / 11.15$  & $6.35 / 12.35$ \\
        CNN                   & $14.78 / 12.69$ & $13.80 / 11.30$ & $18.25 / 14.60$ & $14.55 / 11.95$ & $13.35 / 13.00$ & $15.40 / 13.30$ & $14.35 / 11.80$ & $13.75 / 12.85$ \\
        MXGNet                & $20.78 / 13.07$ & $12.95 / 13.65$ & $37.05 / 13.95$ & $24.80 / 12.50$ & $17.45 / 12.50$ & $16.80 / 12.05$ & $18.05 / 12.95$ & $18.35 / 13.90$ \\
        ResNet                & $24.79 / 13.19$ & $24.30 / 14.50$ & $25.05 / 14.30$ & $25.80 / 12.95$ & $23.80 / 12.35$ & $27.40 / 13.55$ & $25.05 / 13.40$ & $22.15 / 11.30$ \\
        ResNet+DRT            & $31.56 / 13.26$ & $31.65 / 13.20$ & $39.55 / 14.30$ & $35.55 / 13.25$ & $25.65 / 12.15$ & $32.05 / 13.10$ & $31.40 / 13.70$ & $25.05 / 13.15$ \\
        SRAN                  & $15.56 / 29.06$ & $18.35 / 37.55$ & $38.80 / 38.30$ & $17.40 / 29.30$ & $9.45 / 29.55$  & $11.35 / 28.65$ & $5.50 / 21.15$  & $8.05 / 18.95$ \\
        CoPINet               & $52.96 / 22.84$ & $49.45 / 24.50$ & $61.55 / 31.10$ & $\mathbf{52.15} / 25.35$ & $68.10 / 20.60$ & $65.40 / 19.85$ & $39.55 / 19.00$ & $34.55 / 19.45$ \\ 
        \ac{prae} Learner     & $\mathbf{65.03 / 77.02}$ & $\mathbf{76.50 / 90.45}$ & $\mathbf{78.60 / 85.35}$ & $28.55 / \mathbf{45.60}$ & $\mathbf{90.05 / 96.25}$ & $\mathbf{90.85 / 97.35}$ & $\mathbf{48.05 / 63.45}$ & $\mathbf{42.60 / 60.70}$ \\
        \midrule
        Human                 & $84.41$ & $95.45$ & $81.82$ & $79.55$ & $86.36$ & $81.81$ & $86.36$ & $81.81$ \\
        \bottomrule
    \end{tabular}
    }
    \caption{Model performance ($\%$) on RAVEN / I-RAVEN. All models are trained on 2x2Grid only. Acc denotes the mean accuracy. Following Zhang \etal~\cite{zhang2019raven}, L-R is short for the Left-Right configuration, U-D Up-Down, O-IC Out-InCenter, and O-IG Out-InGrid.}
    \label{tbl:gen}
\end{table*}

\begin{table*}[t!]
    \centering
    \resizebox{\linewidth}{!}{
    \begin{tabular}{l c c c c c c c c}
        \toprule
        Object Attribute      & Acc             & Center          & 2x2Grid         & 3x3Grid         & L-R             & U-D             & O-IC            & O-IG           \\
        \midrule
        Objectiveness         & $93.81 / 95.41$ & $96.13 / 96.07$ & $99.79 / 99.99$ & $99.71 / 97.98$ & $99.56 / 95.00$ & $99.86 / 94.84$ & $71.73 / 88.05$ & $82.07 / 95.97$ \\
        Type                  & $86.29 / 89.24$ & $89.89 / 89.33$ & $99.95 / 95.93$ & $83.49 / 85.96$ & $99.92 / 92.90$ & $99.85 / 97.84$ & $91.55 / 91.86$ & $66.68 / 70.85$ \\
        Size                  & $64.72 / 66.63$ & $68.45 / 69.11$ & $71.26 / 73.20$ & $71.42 / 62.02$ & $73.00 / 85.08$ & $73.41 / 73.45$ & $53.54 / 62.63$ & $44.36 / 40.95$ \\
        Color                 & $75.26 / 79.45$ & $75.15 / 75.65$ & $85.15 / 87.81$ & $62.69 / 69.94$ & $85.27 / 83.24$ & $84.45 / 81.38$ & $84.91 / 75.32$ & $78.48 / 82.84$ \\
        \bottomrule
    \end{tabular}
    }
    \caption{Accuracy ($\%$) of the object CNN on each attribute, reported as RAVEN / I-RAVEN. The CNN module is trained with the \ac{prae} learner on 2x2Grid only without any visual attribute annotations. Acc denotes the mean accuracy on each attribute.}
    \label{tbl:attr}
\end{table*}

\begin{table*}[t!]
    \centering
    \resizebox{\linewidth}{!}{
        \begin{tabular}{l c c c c c c c c}
            \toprule
            Panel Attribute       & Acc             & Center           & 2x2Grid         & 3x3Grid         & L-R              & U-D               & O-IC              & O-IG            \\
            \midrule
            Pos/Num               & $90.53 / 91.67$ & -                & $90.55 / 90.05$ & $92.80 / 94.10$ & -                & -                 & -                 & $88.25 / 90.85$ \\
            Type                  & $94.17 / 92.15$ & $100.00 / 95.00$ & $99.75 / 95.30$ & $63.95 / 68.40$ & $100.00 / 99.90$ & $100.00 / 100.00$ & $100.00 / 100.00$ & $86.08 / 77.60$ \\
            Size                  & $90.06 / 88.33$ & $98.95 / 99.00$  & $90.45 / 89.90$ & $65.30 / 70.45$ & $98.15 / 96.78$  & $99.45 / 92.45$   & $93.08 / 96.13$   & $77.35 / 70.78$ \\
            Color                 & $87.38 / 87.25$ & $97.60 / 93.75$  & $88.10 / 85.35$ & $37.45 / 45.65$ & $98.90 / 92.38$  & $99.40 / 98.43$   & $92.90 / 97.23$   & $73.75 / 79.48$ \\
            \bottomrule
        \end{tabular}
    }
    \caption{Accuracy ($\%$) of the probabilistic abduction engine on each attribute, reported as RAVEN / I-RAVEN. The \ac{prae} learner is trained on 2x2Grid only. Acc denotes the mean accuracy on each attribute.}
    \label{tbl:rule}
\end{table*}

\section{Experiments}\label{sec:experiments}

We demonstrate the efficacy of the proposed \ac{prae} learner in \ac{rpm}. In particular, we show that the \ac{prae} learner achieves the best performance among all baselines in the cross-configuration generalization task of \ac{rpm}. In addition, the modularized perception and reasoning process allows us to probe into how each module performs in the \ac{rpm} task and analyze the \ac{prae} learner's strengths and weaknesses. Furthermore, we show that probabilistic scene representation learned by the \ac{prae} learner can be used to generate an answer when equipped with a rendering engine.

\subsection{Experimental Setup}

We evaluate the proposed \ac{prae} learner on RAVEN~\cite{zhang2019raven} and I-RAVEN~\cite{hu2021stratified}. Both datasets consist of $7$ distinct \ac{rpm} configurations, each of which contains $10,000$ samples, equally divided into $6$ folds for training, $2$ folds for validation, and $2$ folds for testing. We compare our \ac{prae} learner with simple baselines of LSTM, CNN, and ResNet, and strong baselines of \ac{wren}~\cite{santoro2018measuring}, ResNet+DRT~\cite{zhang2019raven}, LEN~\cite{zheng2019abstract}, CoPINet~\cite{zhang2019learning}, MXGNet~\cite{wang2020abstract}, and SRAN~\cite{hu2021stratified}. To measure cross-configuration generalization, we train all models using the 2x2Grid configuration due to its proper complexity for probability marginalization and a sufficient number of rules on each panel attribute. We test the models on all \emph{other} configurations. All models are implemented in PyTorch~\cite{paszke2017automatic} and optimized using ADAM~\cite{kingma2014adam} on an Nvidia Titan Xp GPU. For numerical stability, we use log probability in \ac{prae}.

\subsection{Cross-Configuration Generalization}

\cref{tbl:gen} shows the cross-configuration generalization performance of different models. While advanced models like \ac{wren}, LEN, MXGNet, and SRAN have fairly good fitting performance on the training regime, these models fail to learn transferable representation for other configurations, which suggests that they do not learn logics or any forms of abstraction but visual appearance only. Simpler baselines like LSTM, CNNs, ResNet, and ResNet+DRT show less severe overfitting, but neither do they demonstrate satisfactory performance. This effect indicates that using only deep models in abstract visual reasoning makes it very difficult to acquire the generalization capability required in situations with similar inner mechanisms but distinctive appearances. By leveraging the notion of contrast, CoPINet improves generalization performance by a notable margin. 

Equipped with symbolic reasoning and neural perception, not only does the \ac{prae} learner achieve the best performance among all models, but it also shows performance better than humans on three configurations. Compared to baselines trained on the full dataset (see supplementary material), the \ac{prae} learner surpasses all other models on the 2x2Grid domain, despite other models seeing $6$ times more data. The \ac{prae} learner does not exhibit strong overfitting either, achieving comparable and sometimes better performance on Center, L-R, and U-D. However, limitations of the \ac{prae} learner do exist. In cases with overlap (O-IC and O-IG), the performance decreases, and a devastating result is observed on 3x3Grid. The first failure is due to the domain shift in the region appearance that neural models cannot handle, and the second could be attributed to marginalization over probability distributions of multiple objects in 3x3Grid, where uncertainties from all objects accumulate, leading to inaccurate abduced rule distributions. These observations are echoed in our analysis shown next.

\subsection{Analysis on Perception and Reasoning}

RAVEN and I-RAVEN provide multiple levels of annotations for us to analyze our modularized \ac{prae} learner. Specifically, we use the region-based attribute annotations to evaluate our object CNN in perception. Note that the object CNN is not trained using any attribute annotations. We also use the ground-truth rule annotations to evaluate the accuracy of the probabilistic abduction engine. 

\cref{tbl:attr} details the analysis of perception using the object CNN: It achieves reasonable performance on object attribute prediction, though not trained with any visual attribute annotations. The model shows a relatively accurate prediction of objectiveness in order to solve an \ac{rpm} instance. Compared to the size prediction accuracy, the object CNN is better at predicting texture-related attributes of type and color. The object CNN has similar results on 2x2Grid, L-R, and U-D. However, referencing \cref{tbl:gen}, we notice that 2x2Grid requires marginalization over more objects, resulting in an inferior performance. Accuracy further drops on configurations with overlap, leading to unsatisfactory results on O-IC and O-IG. For 3x3Grid, more accurate predictions are necessary as uncertainties accumulate from probabilities over multiple objects.

\cref{tbl:rule} details the analysis on reasoning, showing how the probabilistic abduction engine performs on rule prediction for each attribute across different configurations. Since rules on position and number are exclusive, we merge their performance as Pos/Num. As Center, L-R, U-D, and O-IC do not involve rules on Pos/Num, we do not measure the abduction performance on them. We note that, in general, the abduction engine shows good performance on all panel attributes, with a perfect prediction on type in certain configurations. However, the design of abduction as probability marginalization is a double-edged sword. While the object CNN's performance on size prediction is only marginally different on 2x2Grid and 3x3Grid in RAVEN, their abduction accuracies drastically vary. The difference occurs because uncertainties on object attributes accumulate during marginalization as the number of objects increases, eventually leading to poor performance on rule prediction and answer selection. However, on configurations with fewer objects, unsatisfactory object attribute predictions can still produce accurate rule predictions. Note there is no guarantee that a correct rule will necessarily lead to a correct final choice, as the selected rule still operates on panel attribute distributions inferred from object attribute distributions.

\subsection{Generation Ability}

\begin{figure}[t!]
    \centering
    \includegraphics[width=\linewidth]{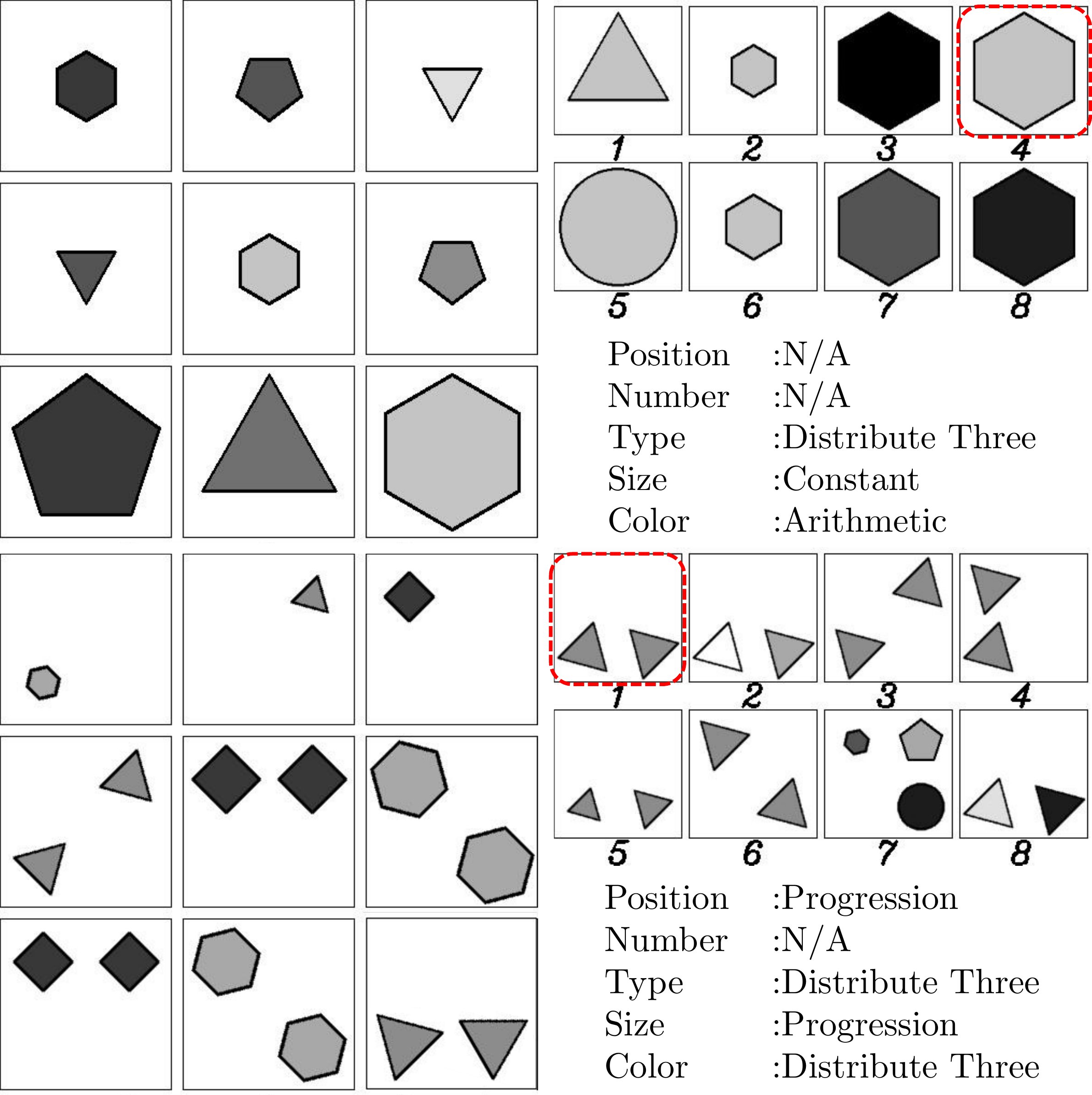}
    \caption{Two \ac{rpm} instances with the final 9th panels filled by our generation results. The ground-truth selections are highlighted in red squares, and the ground-truth rules in each instance are listed. There are no rules on position and number in the first instance of the Center configuration, and the rules on position and number are exclusive in the second instance of 2x2Grid.}
    \label{fig:generate}
\end{figure}

One unique property of the proposed \ac{prae} learner is its ability to directly generate a panel from the predicted representation when a rendering engine is given. The ability resembles the bi-directional top-down and bottom-up reasoning, adding a generative flavor commonly ignored in prior discriminative-only approaches~\cite{hill2019learning,hu2021stratified,santoro2018measuring,wang2020abstract,zhang2019raven,zhang2019learning,zheng2019abstract}. As the \ac{prae} learner predicts final panel attribute distributions and is trained in an analysis-by-synthesis manner, we can sample panel attribute values from the predicted distributions and render the final answer using a rendering engine. Here, we use the rendering program released with RAVEN~\cite{zhang2019raven} to show the generation ability of the \ac{prae} learner. \cref{fig:generate} shows examples of the generation results. Note that one of our generations is slightly different from the ground-truth answer due to random sampling of rotations during rendering. However, it still follows the rules in the problem and should be considered as a correct answer.

\section{Conclusion and Discussion}\label{sec:conclusion}

We propose the \emph{\acf{prae}} learner for spatial-temporal reasoning in \acf{rpm} that decomposes the problem-solving process into neural perception and logical reasoning. While existing methods on \ac{rpm} are merely discriminative, the proposed \ac{prae} learner is a hybrid of generative models and discriminative models, closing the loop in a human-like, top-down bottom-up bi-directional reasoning process. In the experiments, we show that the \ac{prae} learner achieves the best performance on the cross-configuration generalization task on RAVEN and I-RAVEN. The modularized design of the \ac{prae} learner also permits us to probe into how perception and reasoning work independently during problem-solving. Finally, we show the unique generative property of the \ac{prae} learner by filling in the missing panel with an image produced by the values sampled from the probabilistic scene representation.

However, the proposed \ac{prae} learner also has limits. As shown in our experiments, probabilistic abduction can be a double-edged sword in the sense that when the number of objects increases, uncertainties over multiple objects will accumulate, making the entire process sensitive to perception performance. Also, complete probability marginalization introduces a challenge for computational scalability; it prevents us from training the \ac{prae} learner on more complex configurations such as 3x3Grid. One possible solution might be a discrete abduction process. However, jointly learning such a system is non-trivial. It is also difficult for the learner to perceive and reason based on lower-level primitives, such as lines and corners. While, in theory, a generic detector of lines and corners should be able to resolve this issue, no well-performing systems exist in practice, except those with strict handcrafted detection rules, which would miss the critical probabilistic interpretations in the entire framework. The \ac{prae} learner also requires strong prior knowledge about the underlying logical relations to work, while an ideal method should be able to induce the hidden rules by itself. Though a precise induction mechanism is still unknown for humans, an emerging computational technique of bi-level optimization~\cite{finn2017model,zhang2019metastyle} may be able to house perception and induction together into a general optimization framework.

While we answer questions about generalization and generation in \ac{rpm}, one crucial question remains to be addressed: How perception learned from other domains can be transferred and used to solve this abstract reasoning task. Unlike humans that arguably apply knowledge learned from elsewhere to solve \ac{rpm}, current systems still need training on the same task to acquire the capability. While feature transfer is still challenging for computer vision, we anticipate that progress in answering transferability in \ac{rpm} will help address similar questions~\cite{zhang2021acre,zhang2020machine,zhu2020dark} and further advance the field.

\textbf{Acknowledgement:}
The authors thank Sirui Xie, Prof. Ying Nian Wu, and Prof. Hongjing Lu at UCLA for helpful discussions. The work reported herein was supported by ONR MURI grant N00014-16-1-2007, DARPA XAI grant N66001-17-2-4029, and ONR grant N00014-19-1-2153.

{\small
\bibliographystyle{ieee_fullname}
\bibliography{bib}

\begin{thebibliography}{10}\itemsep=-1pt

\bibitem{antol2015vqa}
Stanislaw Antol, Aishwarya Agrawal, Jiasen Lu, Margaret Mitchell, Dhruv Batra,
  C Lawrence~Zitnick, and Devi Parikh.
\newblock Vqa: Visual question answering.
\newblock In {\em Proceedings of International Conference on Computer Vision
  (ICCV)}, 2015.

\bibitem{bengio2009curriculum}
Yoshua Bengio, J{\'e}r{\^o}me Louradour, Ronan Collobert, and Jason Weston.
\newblock Curriculum learning.
\newblock In {\em Proceedings of International Conference on Machine Learning
  (ICML)}, 2009.

\bibitem{carpenter1990one}
Patricia~A Carpenter, Marcel~A Just, and Peter Shell.
\newblock What one intelligence test measures: a theoretical account of the
  processing in the raven progressive matrices test.
\newblock {\em Psychological Review}, 97(3):404, 1990.

\bibitem{chen2019holistic}
Yixin Chen, Siyuan Huang, Tao Yuan, Yixin Zhu, Siyuan Qi, and Song-Chun Zhu.
\newblock Holistic++ scene understanding: Single-view 3d holistic scene parsing
  and human pose estimation with human-object interaction and physical
  commonsense.
\newblock In {\em Proceedings of International Conference on Computer Vision
  (ICCV)}, 2019.

\bibitem{dehaene2006core}
Stanislas Dehaene, V{\'e}ronique Izard, Pierre Pica, and Elizabeth Spelke.
\newblock Core knowledge of geometry in an amazonian indigene group.
\newblock {\em Science}, 311(5759):381--384, 2006.

\bibitem{snow1984the}
R E~Snow, Patrick Kyllonen, and B Marshalek.
\newblock The topography of ability and learning correlations.
\newblock {\em Advances in the psychology of human intelligence}, pages
  47--103, 1984.

\bibitem{edmonds2018human}
Mark Edmonds, Feng Kubricht, James, Colin Summers, Yixin Zhu, Brandon Rothrock,
  Song-Chun Zhu, and Hongjing Lu.
\newblock Human causal transfer: Challenges for deep reinforcement learning.
\newblock In {\em Proceedings of the Annual Meeting of the Cognitive Science
  Society (CogSci)}, 2018.

\bibitem{edmonds2020theory}
Mark Edmonds, Xiaojian Ma, Siyuan Qi, Yixin Zhu, Hongjing Lu, and Song-Chun
  Zhu.
\newblock Theory-based causal transfer: Integrating instance-level induction
  and abstract-level structure learning.
\newblock In {\em Proceedings of AAAI Conference on Artificial Intelligence
  (AAAI)}, 2020.

\bibitem{edmonds2019decomposing}
Mark Edmonds, Siyuan Qi, Yixin Zhu, James Kubricht, Song-Chun Zhu, and Hongjing
  Lu.
\newblock Decomposing human causal learning: Bottom-up associative learning and
  top-down schema reasoning.
\newblock In {\em Proceedings of the Annual Meeting of the Cognitive Science
  Society (CogSci)}, 2019.

\bibitem{finn2017model}
Chelsea Finn, Pieter Abbeel, and Sergey Levine.
\newblock Model-agnostic meta-learning for fast adaptation of deep networks.
\newblock In {\em Proceedings of International Conference on Machine Learning
  (ICML)}, 2017.

\bibitem{fodor1988connectionism}
Jerry~A Fodor, Zenon~W Pylyshyn, et~al.
\newblock Connectionism and cognitive architecture: A critical analysis.
\newblock {\em Cognition}, 28(1-2):3--71, 1988.

\bibitem{ghallab2004automated}
Malik Ghallab, Dana Nau, and Paolo Traverso.
\newblock {\em Automated Planning: theory and practice}.
\newblock Elsevier, 2004.

\bibitem{grandin2006thinking}
Temple Grandin.
\newblock {\em Thinking in pictures: And other reports from my life with
  autism}.
\newblock Vintage, 2006.

\bibitem{grenander1976lectures}
Ulf Grenander.
\newblock Lectures in pattern theory i, ii and iii: Pattern analysis, pattern
  synthesis and regular structures, 1976.

\bibitem{han2019visual}
Chi Han, Jiayuan Mao, Chuang Gan, Josh Tenenbaum, and Jiajun Wu.
\newblock Visual concept-metaconcept learning.
\newblock In {\em Proceedings of Advances in Neural Information Processing
  Systems (NeurIPS)}, 2019.

\bibitem{han2019divergence}
Tian Han, Erik Nijkamp, Xiaolin Fang, Mitch Hill, Song-Chun Zhu, and Ying~Nian
  Wu.
\newblock Divergence triangle for joint training of generator model,
  energy-based model, and inferential model.
\newblock In {\em Proceedings of the IEEE Conference on Computer Vision and
  Pattern Recognition (CVPR)}, 2019.

\bibitem{hill2019learning}
Felix Hill, Adam Santoro, David~GT Barrett, Ari~S Morcos, and Timothy
  Lillicrap.
\newblock Learning to make analogies by contrasting abstract relational
  structure.
\newblock In {\em International Conference on Learning Representations (ICLR)},
  2019.

\bibitem{hofstadter1995fluid}
Douglas~R Hofstadter.
\newblock {\em Fluid concepts and creative analogies: Computer models of the
  fundamental mechanisms of thought.}
\newblock Basic books, 1995.

\bibitem{holyoak2012oxford}
Keith~James Holyoak and Robert~G Morrison.
\newblock {\em The Oxford handbook of thinking and reasoning}.
\newblock Oxford University Press, 2012.

\bibitem{hu2021stratified}
Sheng Hu, Yuqing Ma, Xianglong Liu, Yanlu Wei, and Shihao Bai.
\newblock Stratified rule-aware network for abstract visual reasoning.
\newblock In {\em Proceedings of AAAI Conference on Artificial Intelligence
  (AAAI)}, 2021.

\bibitem{huang2019continuous}
De-An Huang, Danfei Xu, Yuke Zhu, Animesh Garg, Silvio Savarese, Li Fei-Fei,
  and Juan~Carlos Niebles.
\newblock Continuous relaxation of symbolic planner for one-shot imitation
  learning.
\newblock In {\em Proceedings of International Conference on Intelligent Robots
  and Systems (IROS)}, 2019.

\bibitem{huang2018cooperative}
Siyuan Huang, Siyuan Qi, Yinxue Xiao, Yixin Zhu, Ying~Nian Wu, and Song-Chun
  Zhu.
\newblock Cooperative holistic scene understanding: Unifying 3d object, layout
  and camera pose estimation.
\newblock In {\em Proceedings of Advances in Neural Information Processing
  Systems (NeurIPS)}, 2018.

\bibitem{huang2018holistic}
Siyuan Huang, Siyuan Qi, Yixin Zhu, Yinxue Xiao, Yuanlu Xu, and Song-Chun Zhu.
\newblock Holistic 3d scene parsing and reconstruction from a single rgb image.
\newblock In {\em Proceedings of European Conference on Computer Vision
  (ECCV)}, 2018.

\bibitem{hunt1974quote}
Earl Hunt.
\newblock {\em Quote the Raven? Nevermore.}
\newblock Lawrence Erlbaum, 1974.

\bibitem{izard2011flexible}
V{\'e}ronique Izard, Pierre Pica, Elizabeth~S Spelke, and Stanislas Dehaene.
\newblock Flexible intuitions of euclidean geometry in an amazonian indigene
  group.
\newblock {\em Proceedings of the National Academy of Sciences (PNAS)},
  108(24):9782--9787, 2011.

\bibitem{jaeggi2008improving}
Susanne~M Jaeggi, Martin Buschkuehl, John Jonides, and Walter~J Perrig.
\newblock Improving fluid intelligence with training on working memory.
\newblock {\em Proceedings of the National Academy of Sciences (PNAS)},
  105(19):6829--6833, 2008.

\bibitem{jang2017tgif}
Yunseok Jang, Yale Song, Youngjae Yu, Youngjin Kim, and Gunhee Kim.
\newblock Tgif-qa: Toward spatio-temporal reasoning in visual question
  answering.
\newblock In {\em Proceedings of the IEEE Conference on Computer Vision and
  Pattern Recognition (CVPR)}, 2017.

\bibitem{jordan1992forward}
Michael~I Jordan and David~E Rumelhart.
\newblock Forward models: Supervised learning with a distal teacher.
\newblock {\em Cognitive Science}, 16(3):307--354, 1992.

\bibitem{kansky2017schema}
Ken Kansky, Tom Silver, David~A M{\'e}ly, Mohamed Eldawy, Miguel
  L{\'a}zaro-Gredilla, Xinghua Lou, Nimrod Dorfman, Szymon Sidor, Scott
  Phoenix, and Dileep George.
\newblock Schema networks: Zero-shot transfer with a generative causal model of
  intuitive physics.
\newblock In {\em Proceedings of International Conference on Machine Learning
  (ICML)}, 2017.

\bibitem{kingma2014adam}
Diederik~P Kingma and Jimmy Ba.
\newblock Adam: A method for stochastic optimization.
\newblock In {\em International Conference on Learning Representations (ICLR)},
  2014.

\bibitem{konidaris2015symbol}
George Konidaris, Leslie Kaelbling, and Tomas Lozano-Perez.
\newblock Symbol acquisition for probabilistic high-level planning.
\newblock In {\em Proceedings of International Joint Conference on Artificial
  Intelligence (IJCAI)}, 2015.

\bibitem{lecun1998gradient}
Yann LeCun, L{\'e}on Bottou, Yoshua Bengio, Patrick Haffner, et~al.
\newblock Gradient-based learning applied to document recognition.
\newblock {\em Proceedings of the IEEE}, 86(11):2278--2324, 1998.

\bibitem{li2020closed}
Qing Li, Siyuan Huang, Yining Hong, Yixin Chen, Ying~Nian Wu, and Song-Chun
  Zhu.
\newblock Closed loop neural-symbolic learning via integrating neural
  perception, grammar parsing, and symbolic reasoning.
\newblock In {\em Proceedings of International Conference on Machine Learning
  (ICML)}, 2020.

\bibitem{lin1991divergence}
Jianhua Lin.
\newblock Divergence measures based on the shannon entropy.
\newblock {\em IEEE Transactions on Information theory}, 37(1):145--151, 1991.

\bibitem{little2012bayesian}
Daniel~R Little, Stephan Lewandowsky, and Thomas~L Griffiths.
\newblock A bayesian model of rule induction in raven's progressive matrices.
\newblock In {\em Proceedings of the Annual Meeting of the Cognitive Science
  Society (CogSci)}, 2012.

\bibitem{loper2014opendr}
Matthew~M Loper and Michael~J Black.
\newblock Opendr: An approximate differentiable renderer.
\newblock In {\em Proceedings of European Conference on Computer Vision
  (ECCV)}, 2014.

\bibitem{lovett2017modeling}
Andrew Lovett and Kenneth Forbus.
\newblock Modeling visual problem solving as analogical reasoning.
\newblock {\em Psychological Review}, 124(1):60, 2017.

\bibitem{lovett2010structure}
Andrew Lovett, Kenneth Forbus, and Jeffrey Usher.
\newblock A structure-mapping model of raven's progressive matrices.
\newblock In {\em Proceedings of the Annual Meeting of the Cognitive Science
  Society (CogSci)}, 2010.

\bibitem{lovett2009solving}
Andrew Lovett, Emmett Tomai, Kenneth Forbus, and Jeffrey Usher.
\newblock Solving geometric analogy problems through two-stage analogical
  mapping.
\newblock {\em Cognitive Science}, 33(7):1192--1231, 2009.

\bibitem{mao2019neuro}
Jiayuan Mao, Chuang Gan, Pushmeet Kohli, Joshua~B Tenenbaum, and Jiajun Wu.
\newblock The neuro-symbolic concept learner: Interpreting scenes, words, and
  sentences from natural supervision.
\newblock In {\em International Conference on Learning Representations (ICLR)},
  2019.

\bibitem{marcus2019rebooting}
Gary Marcus and Ernest Davis.
\newblock {\em Rebooting AI: building artificial intelligence we can trust}.
\newblock Pantheon, 2019.

\bibitem{marcus2020insights}
Gary Marcus and Ernest Davis.
\newblock Insights for ai from the human mind.
\newblock {\em Communications of the ACM}, 64(1):38--41, 2020.

\bibitem{marcus1998rethinking}
Gary~F Marcus.
\newblock Rethinking eliminative connectionism.
\newblock {\em Cognitive psychology}, 37(3):243--282, 1998.

\bibitem{marcus2018algebraic}
Gary~F Marcus.
\newblock {\em The algebraic mind: Integrating connectionism and cognitive
  science}.
\newblock MIT press, 2018.

\bibitem{mcgreggor2014confident}
Keith McGreggor and Ashok Goel.
\newblock Confident reasoning on raven's progressive matrices tests.
\newblock In {\em Proceedings of AAAI Conference on Artificial Intelligence
  (AAAI)}, 2014.

\bibitem{mcgreggor2014fractals}
Keith McGreggor, Maithilee Kunda, and Ashok Goel.
\newblock Fractals and ravens.
\newblock {\em Artificial Intelligence}, 215:1--23, 2014.

\bibitem{mekik2018similarity}
Can~Serif Mekik, Ron Sun, and David~Yun Dai.
\newblock Similarity-based reasoning, raven's matrices, and general
  intelligence.
\newblock In {\em Proceedings of International Joint Conference on Artificial
  Intelligence (IJCAI)}, 2018.

\bibitem{paszke2017automatic}
Adam Paszke, Sam Gross, Soumith Chintala, Gregory Chanan, Edward Yang, Zachary
  DeVito, Zeming Lin, Alban Desmaison, Luca Antiga, and Adam Lerer.
\newblock Automatic differentiation in {PyTorch}.
\newblock In {\em NIPS Autodiff Workshop}, 2017.

\bibitem{qi2020generalized}
Siyuan Qi, Baoxiong Jia, Siyuan Huang, Ping Wei, and Song-Chun Zhu.
\newblock A generalized earley parser for human activity parsing and
  prediction.
\newblock {\em IEEE Transactions on Pattern Analysis and Machine Intelligence
  (TPAMI)}, 2020.

\bibitem{qi2018generalized}
Siyuan Qi, Baoxiong Jia, and Song-Chun Zhu.
\newblock Generalized earley parser: Bridging symbolic grammars and sequence
  data for future prediction.
\newblock In {\em Proceedings of International Conference on Machine Learning
  (ICML)}, 2018.

\bibitem{raven1936mental}
James~C Raven.
\newblock Mental tests used in genetic studies: The performance of related
  individuals on tests mainly educative and mainly reproductive.
\newblock Master's thesis, University of London, 1936.

\bibitem{raven1998raven}
John~C Raven and John~Hugh Court.
\newblock {\em Raven's progressive matrices and vocabulary scales}.
\newblock Oxford pyschologists Press, 1998.

\bibitem{santoro2018measuring}
Adam Santoro, Felix Hill, David Barrett, Ari Morcos, and Timothy Lillicrap.
\newblock Measuring abstract reasoning in neural networks.
\newblock In {\em Proceedings of International Conference on Machine Learning
  (ICML)}, 2018.

\bibitem{shegheva2018structural}
Snejana Shegheva and Ashok Goel.
\newblock The structural affinity method for solving the raven's progressive
  matrices test for intelligence.
\newblock In {\em Proceedings of AAAI Conference on Artificial Intelligence
  (AAAI)}, 2018.

\bibitem{spearman1927abilities}
Charles Spearman.
\newblock {\em The abilities of man}.
\newblock Macmillan, 1927.

\bibitem{spratley2020closer}
Steven Spratley, Krista Ehinger, and Tim Miller.
\newblock A closer look at generalisation in raven.
\newblock In {\em Proceedings of European Conference on Computer Vision
  (ECCV)}, 2020.

\bibitem{steenbrugge2018improving}
Xander Steenbrugge, Sam Leroux, Tim Verbelen, and Bart Dhoedt.
\newblock Improving generalization for abstract reasoning tasks using
  disentangled feature representations.
\newblock {\em arXiv preprint arXiv:1811.04784}, 2018.

\bibitem{sutton1998introduction}
Richard~S Sutton, Andrew~G Barto, et~al.
\newblock {\em Introduction to reinforcement learning}.
\newblock MIT press Cambridge, 1998.

\bibitem{wang2020abstract}
Duo Wang, Mateja Jamnik, and Pietro Lio.
\newblock Abstract diagrammatic reasoning with multiplex graph networks.
\newblock In {\em International Conference on Learning Representations (ICLR)},
  2020.

\bibitem{wang2015automatic}
Ke Wang and Zhendong Su.
\newblock Automatic generation of raven’s progressive matrices.
\newblock In {\em Proceedings of International Joint Conference on Artificial
  Intelligence (IJCAI)}, 2015.

\bibitem{williams1992simple}
Ronald~J Williams.
\newblock Simple statistical gradient-following algorithms for connectionist
  reinforcement learning.
\newblock {\em Machine learning}, 8(3-4):229--256, 1992.

\bibitem{wu2017neural}
Jiajun Wu, Joshua~B Tenenbaum, and Pushmeet Kohli.
\newblock Neural scene de-rendering.
\newblock In {\em Proceedings of the IEEE Conference on Computer Vision and
  Pattern Recognition (CVPR)}, 2017.

\bibitem{wu2017marrnet}
Jiajun Wu, Yifan Wang, Tianfan Xue, Xingyuan Sun, Bill Freeman, and Josh
  Tenenbaum.
\newblock Marrnet: 3d shape reconstruction via 2.5 d sketches.
\newblock In {\em Proceedings of Advances in Neural Information Processing
  Systems (NeurIPS)}, 2017.

\bibitem{xie2016theory}
Jianwen Xie, Yang Lu, Song-Chun Zhu, and Yingnian Wu.
\newblock A theory of generative convnet.
\newblock In {\em Proceedings of International Conference on Machine Learning
  (ICML)}, 2016.

\bibitem{xie2019learning}
Jianwen Xie, Song-Chun Zhu, and Ying~Nian Wu.
\newblock Learning energy-based spatial-temporal generative convnets for
  dynamic patterns.
\newblock {\em IEEE Transactions on Pattern Analysis and Machine Intelligence
  (TPAMI)}, 2019.

\bibitem{xie2021halma}
Sirui Xie, Xiaojian Ma, Peiyu Yu, Yixin Zhu, Ying~Nian Wu, and Song-Chun Zhu.
\newblock Halma: Humanlike abstraction learning meets affordance in rapid
  problem solving.
\newblock {\em arXiv preprint arXiv:2102.11344}, 2021.

\bibitem{yi2020clevrer}
Kexin Yi, Chuang Gan, Yunzhu Li, Pushmeet Kohli, Jiajun Wu, Antonio Torralba,
  and Joshua Tenenbaum.
\newblock Clevrer: Collision events for video representation and reasoning.
\newblock In {\em International Conference on Learning Representations (ICLR)},
  2020.

\bibitem{yi2018neural}
Kexin Yi, Jiajun Wu, Chuang Gan, Antonio Torralba, Pushmeet Kohli, and Josh
  Tenenbaum.
\newblock Neural-symbolic vqa: Disentangling reasoning from vision and language
  understanding.
\newblock In {\em Proceedings of Advances in Neural Information Processing
  Systems (NeurIPS)}, 2018.

\bibitem{yuille2006vision}
Alan Yuille and Daniel Kersten.
\newblock Vision as bayesian inference: analysis by synthesis?
\newblock {\em Trends in cognitive sciences}, 2006.

\bibitem{zhang2019raven}
Chi Zhang, Feng Gao, Baoxiong Jia, Yixin Zhu, and Song-Chun Zhu.
\newblock Raven: A dataset for relational and analogical visual reasoning.
\newblock In {\em Proceedings of the IEEE Conference on Computer Vision and
  Pattern Recognition (CVPR)}, 2019.

\bibitem{zhang2021acre}
Chi Zhang, Baoxiong Jia, Mark Edmonds, Song-Chun Zhu, and Yixin Zhu.
\newblock Acre: Abstract causal reasoning beyond covariation.
\newblock In {\em Proceedings of the IEEE Conference on Computer Vision and
  Pattern Recognition (CVPR)}, 2021.

\bibitem{zhang2019learning}
Chi Zhang, Baoxiong Jia, Feng Gao, Yixin Zhu, Hongjing Lu, and Song-Chun Zhu.
\newblock Learning perceptual inference by contrasting.
\newblock In {\em Proceedings of Advances in Neural Information Processing
  Systems (NeurIPS)}, 2019.

\bibitem{zhang2019metastyle}
Chi Zhang, Yixin Zhu, and Song-Chun Zhu.
\newblock Metastyle: Three-way trade-off among speed, flexibility, and quality
  in neural style transfer.
\newblock In {\em Proceedings of AAAI Conference on Artificial Intelligence
  (AAAI)}, 2019.

\bibitem{zhang2020machine}
Wenhe Zhang, Chi Zhang, Yixin Zhu, and Song-Chun Zhu.
\newblock Machine number sense: A dataset of visual arithmetic problems for
  abstract and relational reasoning.
\newblock In {\em Proceedings of AAAI Conference on Artificial Intelligence
  (AAAI)}, 2020.

\bibitem{zheng2019abstract}
Kecheng Zheng, Zheng-Jun Zha, and Wei Wei.
\newblock Abstract reasoning with distracting features.
\newblock In {\em Proceedings of Advances in Neural Information Processing
  Systems (NeurIPS)}, 2019.

\bibitem{zhu2007stochastic}
Song-Chun Zhu and David Mumford.
\newblock A stochastic grammar of images.
\newblock {\em Foundations and Trends{\textregistered} in Computer Graphics and
  Vision}, 2(4):259--362, 2007.

\bibitem{zhu1998filters}
Song-Chun Zhu, Yingnian Wu, and David Mumford.
\newblock Filters, random fields and maximum entropy (frame): Towards a unified
  theory for texture modeling.
\newblock {\em International Journal of Computer Vision (IJCV)},
  27(2):107--126, 1998.

\bibitem{zhu2020dark}
Yixin Zhu, Tao Gao, Lifeng Fan, Siyuan Huang, Mark Edmonds, Hangxin Liu, Feng
  Gao, Chi Zhang, Siyuan Qi, Ying~Nian Wu, et~al.
\newblock Dark, beyond deep: A paradigm shift to cognitive ai with humanlike
  common sense.
\newblock {\em Engineering}, 6(3):310--345, 2020.

\end{thebibliography}
}

\end{document}